\documentclass[letterpaper, 10 pt, conference]{ieeeconf}  % Comment this line out if you need a4paper

\IEEEoverridecommandlockouts                              % This command is only needed if 
\makeatletter
\let\NAT@parse\undefined
\makeatother
\usepackage[numbers,sort&compress]{natbib}

\usepackage{graphicx} %
\usepackage{subcaption}
\usepackage[usenames,dvipsnames,table]{xcolor}
\usepackage{booktabs}
\usepackage{xspace}
\usepackage{multirow}
\usepackage[most]{tcolorbox}
\usepackage{amssymb}
\usepackage{arydshln}
\usepackage{bbding}
\usepackage{makecell}
\usepackage[linesnumbered,ruled,vlined]{algorithm2e}
\usepackage{algpseudocode}
\usepackage{pbalance}

\usepackage{hyperref}
\hypersetup{
    colorlinks=true,
    linkcolor=MidnightBlue,
    filecolor=magenta,      
    urlcolor=MidnightBlue,
    citecolor=magenta,
}
\usepackage[capitalise, nameinlink]{cleveref}
\newcommand{\eg}{\textit{e.g.},\ }
\newcommand{\ie}{\textit{i.e.},\ }

\newcommand{\framework}{FadeLead\xspace}
\newcommand{\circledblack}[1]{%
    \tikz[baseline=(char.base)]{%
        \node[draw=white, % Optional: white border for the circle
              circle,
              fill=black, % Explicitly fill the circle with black
              inner sep=1pt, % Adjust spacing inside the circle
              text=white, % Explicitly set text color to white
              font=\bfseries % Optional: make the character bold
              ] (char) {#1};%
    }%
}
\newcommand{\circledred}[1]{%
    \tikz[baseline=(char.base)]{%
        \node[draw=white, % Optional: white border for the circle
              circle,
              fill=red, % Explicitly fill the circle with black
              inner sep=1pt, % Adjust spacing inside the circle
              text=white, % Explicitly set text color to white
              font=\bfseries % Optional: make the character bold
              ] (char) {#1};%
    }%
}

\definecolor{edgeblue}{HTML}{61CBF4}
\newcommand{\circledvis}[1][2.2ex]{%
  \tikz[baseline=(c.base), text height=1.2ex, text depth=.25ex]{%
    \node[circle,
          draw=edgeblue,
          dashed,
          line width=1pt,
          fill=none,
          minimum size=#1,
          inner sep=0pt,
          outer sep=0pt] (c) {};%
  }%
}

\definecolor{edgegreen}{HTML}{28A745}
\newcommand{\rectvis}[1][2.2ex]{%
  \tikz[baseline=(r.base), text height=1.2ex, text depth=.25ex]{%
    \node[rectangle,
          draw=edgegreen,
          dashed,
          line width=1pt,
          fill=none,
          minimum width=1.5*#1,
          minimum height=0.6*#1,
          inner sep=0pt,
          outer sep=0pt] (r) {};%
  }%
}

\title{\LARGE \bf \textit{Background \underline{Fade}s, Foreground \underline{Lead}s}: Curriculum-Guided Background Pruning for Efficient Foreground-Centric Collaborative Perception
}

\author{Yuheng Wu$^{1}$, Xiangbo Gao$^{2}$, Quang Tau$^{1}$, Zhengzhong Tu$^{2}$ and Dongman Lee$^{1}$
\thanks{$^{1}$KAIST \{yuhengwu, quangntau1223, dlee\}@kaist.ac.kr}
\thanks{$^{2}$Texas A\&M University \{xiangbog, tzz\}@tamu.edu}
}

\begin{document}
\definecolor{cvprblue}{rgb}{0.21,0.49,0.74}

\maketitle
\thispagestyle{empty}
\pagestyle{empty}

\begin{abstract}
Collaborative perception enhances the reliability and spatial coverage of autonomous vehicles by sharing complementary information across vehicles, offering a promising solution to long-tail scenarios that challenge single-vehicle perception. However, the bandwidth constraints of vehicular networks make transmitting the entire feature map impractical. Recent methods, therefore, adopt a foreground-centric paradigm, transmitting only predicted foreground-region features while discarding the background, which encodes essential context. We propose \framework, a foreground-centric framework that overcomes this limitation by learning to encapsulate background context into compact foreground features during training. At the core of our design is a curricular learning strategy that leverages background cues early on but progressively prunes them away, forcing the model to internalize context into foreground representations without transmitting background itself. Extensive experiments on both simulated and real-world benchmarks show that \framework outperforms prior methods under different bandwidth settings, underscoring the effectiveness of context-enriched foreground sharing. Code, checkpoints and visualizations are availabe at \url{https://github.com/wyhallenwu/FadeLead}.
\end{abstract}
\section{Introduction}
\vspace{-0.2em}
Collaborative perception has emerged as a cornerstone in advancing the safety and reliability of autonomous driving systems. By enabling multiple vehicles to share complementary sensory information, it overcomes critical limitations of single vehicle perception, such as occlusions, restricted fields of view, and blind spots~\cite{gaoSTAMPScalableTask2025,luExtensibleFrameworkOpen2024,gao2025langcoop,huWhere2commCommunicationEfficientCollaborative2022,liuWhen2comMultiAgentPerception2020,yangHow2commCommunicationEfficientCollaborationPragmatic2023,codefilling,wang2025coopdetr,lu2022robust,zhang2025co,yang2024align,shao2024hetecooper,tao2024directed,xu2025codyntrust,huangActFormerScalableCollaborative2024}. Through multi-agent cooperation, collaborative perception systems achieve broader spatial coverage and improved perception robustness.

Despite these benefits, real-world deployment must operate under stringent communication constraints imposed by vehicular networks. A central challenge, therefore, is determining \textit{what subset of features} to transmit to maximize perception utility while minimizing bandwidth consumption.

To address this challenge, recent approaches~\cite{huWhere2commCommunicationEfficientCollaborative2022,xu2025cosdh,yangHow2commCommunicationEfficientCollaborationPragmatic2023} have predominantly adopted a \textit{foreground-centric paradigm}. The intuition is straightforward: object-centric regions (\eg vehicles, pedestrians) are compact in nature, highly task-relevant, and therefore prioritize them for transmission. Typical pipelines learn a confidence map to estimate the foreground likelihood of each BEV cell, and only the top-$k$ high-confidence regions are transmitted. However, this foreground-centric paradigm implicitly treats background as redundant and assumes that confident foreground regions alone are sufficient for robust collaborative perception. This raises a fundamental and underexplored question:

\begin{center}
    \textcolor{cvprblue}{\textit{Is background, which often constitutes the majority of the scene yet is typically discarded, truly redundant?}}
\end{center}

\begin{figure}[t] 
    \centering
    \hspace{-0.5em}
    \includegraphics[width=\columnwidth]{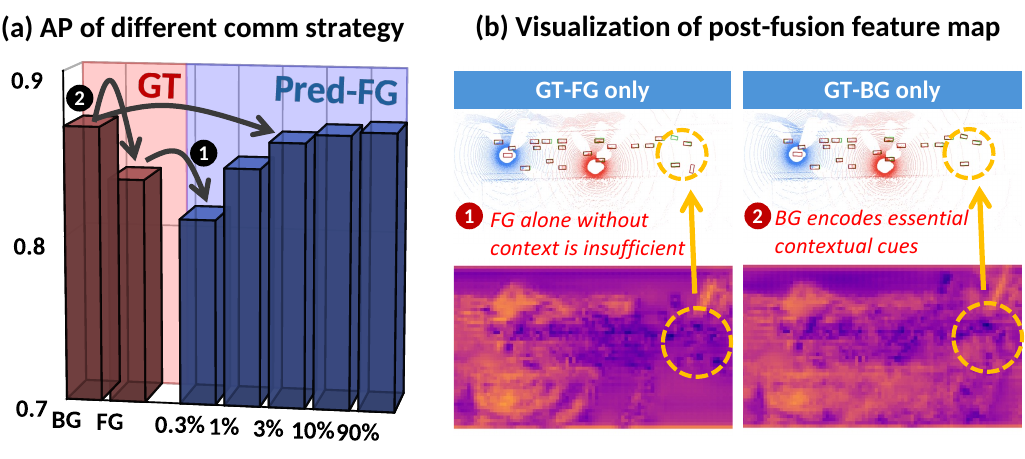}
    \vspace{-2em}
    \caption{(a) Average Precision (AP) of Where2Comm~\cite{huWhere2commCommunicationEfficientCollaborative2022} on OPV2V~\cite{opv2v} under different communication strategies and (b) visualization of the post-fusion feature map.}
    \label{fig:motivation}
\end{figure}

\noindent\textbf{Motivation.} To ground our study, we first revisit the representative method Where2Comm~\cite{huWhere2commCommunicationEfficientCollaborative2022} and disentangle the roles of three feature types: (1) Predicted Foreground (Pred-FG): regions selected by confidence scores. (2) Ground-Truth Foreground (GT-FG): precise object regions from ground-truth annotations. (3) Ground-Truth Background (GT-BG): all remaining regions complementing GT-FG.
We design two experiments to investigate their contribution:

\textbf{Experiment\circledblack{1} $\boldsymbol{\rightarrow}$ \textcolor{red}{Insight}\circledred{1}: GT-FG Only.}
% We replace Pred-FG with GT-FG, transmitting only precise object regions. Despite occupying just \(\sim\)0.3\% (average on test split) of the BEV plane, they outperform the same fraction of Pred-FG, demonstrating that accurate object localization is more reliable. However, performance of Pred-FG continues to improve as the ratio increases (\eg 0.3\%$\rightarrow$90\% \textcolor{blue}{blue bars} in Fig.~\ref{fig:motivation}a), revealing that \textbf{foreground alone, even when perfectly localized, is incomplete} because it misses inter-object dependencies and broader scene semantics.
We replace Pred-FG with GT-FG, transmitting only precise object regions. Although these regions occupy merely \(\sim\)0.3\% (on average across the test split) of the BEV plane, they outperform the same fraction of Pred-FG, indicating that accurate object localization is more reliable. However, as the selection ratio increases (\eg 0.3\%$\rightarrow$90\% \textcolor{blue}{blue bars} in Fig.~\ref{fig:motivation}a), the performance of Pred-FG continues to improve. These results reveal that \textbf{foreground alone, even when perfectly localized, remains insufficient}, as it neglects inter-object dependencies and broader scene semantics.

\textbf{Experiment\circledblack{2} $\boldsymbol{\rightarrow}$ \textcolor{red}{Insight}\circledred{2}: GT-BG Only.}
Conversely, we mask out all GT-FG regions and transmit only the background region features. Surprisingly, as shown in Fig.~\ref{fig:motivation}, the GT-BG strategy not only surpasses the GT-FG counterpart but also achieves performance comparable to sharing nearly the entire BEV feature map. Moreover, as shown in Fig.~\ref{fig:motivation}b, they yield cleaner post-fusion representations. These results demonstrate that \textbf{background is not redundant but carries rich contextual cues} that are critical for disambiguation, robustness under occlusions, and holistic scene understanding.

Together, these insights motivate a new perspective on efficient, bandwidth-constrained collaborative perception:

% \noindent\textbf{Insights.} These experimental results reveal two critical limitations of current foreground-centric strategy: \textbf{\circled{1} Foreground alone is effective but incomplete.} Ground-truth foreground masks demonstrate that precisely localized objects are compact and reliable, outperforming noisy high-confidence predictions. However, as the selection ratio grows, performance gain improves because foreground-only features cannot capture inter-object relations and broader scene semantics. \circled{2} \textbf{Background encodes essential context.} Background regions are not redundant and they carry rich contextual cues critical for disambiguation, robustness to occlusions, and perception completeness. 

\begin{center}
    \textcolor{cvprblue}{\textit{Foreground should not stand alone. It must be enriched by the essential context encoded in the background.}}
\end{center}

\noindent\textbf{Proposed Framework.}  
We propose \framework, a collaborative perception framework that \textit{remains foreground-centric} but addresses the incompleteness of foreground-only sharing. The core idea is to enrich compact foreground representations with critical background context, thereby preserving communication efficiency while enhancing perception robustness. To this end, \framework sharpens shared foreground features through three complementary modules: (1) \textit{Foreground–Context Attention (FCA)} enriches foreground features by querying the full scene for complementary cues. (2) \textit{Curricular Background Pruning (CBP)}, our major contribution, selectively mines informative background during training and progressively prunes it with an annealing schedule, forcing the model to internalize background-derived context into compact foreground features. At inference, only the enriched foreground is transmitted, enabling efficient yet context-aware collaboration. (3) \textit{Foreground Amplification Fusion (FAF)} selectively enriches and amplifies salient ego foreground features by fusing complementary cues from local observations and received neighbor features, ensuring robust perception under strict bandwidth constraints.    

\noindent\textbf{Contributions.} We summarize our contributions as follows:  
\begin{itemize}
    \item We conduct a systematic investigation on the role of background in collaborative perception. Our motivating study reveals that background region features, typically discarded by foreground-centric strategies, encode essential contextual cues for robust perception.  
    \item We propose \framework, a curricular training framework that \textit{remains foreground-centric} but strengthens it by encapsulating informative background context into compact shared foreground features, thereby improving both bandwidth efficiency and contextual completeness.
    \item We demonstrate through extensive experiments on simulated and real-world scenarios that \framework achieves superior detection accuracy under the same bandwidth budget, and remains highly effective even with extremely low selection ratios (\ie 1\%).   
\end{itemize}

\section{Related Work}

\begin{figure}[t!] \centering
    \includegraphics[width=0.49\textwidth]{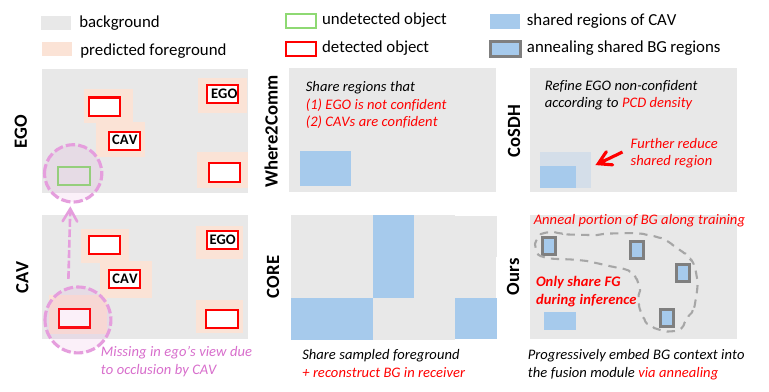}
    \caption{Comparison of different foreground-centric sharing strategy.} \label{fig:comparison}
\end{figure}

\noindent\textbf{Collaborative Perception.} The information-sharing mechanisms in collaborative perception can be broadly categorized into three paradigms: (1) Early fusion~\cite{chenCooperCooperativePerception2019, arnold2020cooperative}, which transmits raw sensor data (\eg images or point clouds) between vehicles. While this approach offers high perception accuracy, it incurs prohibitive communication costs due to the large data volumes of raw inputs. (2) Late fusion~\cite{xu2025cosdh, opv2v}, which exchanges only the final detection results from each vehicle. This approach is bandwidth-efficient but suffers from higher latency (since full-model inference must be completed before sharing) and low update frequency of detection outputs. (3) Intermediate fusion~\cite{liuWhen2comMultiAgentPerception2020, wangV2VNetVehicletoVehicleCommunication2020, xuV2XViTVehicletoEverythingCooperative2022, huWhere2commCommunicationEfficientCollaborative2022, wangCoreCooperativeReconstruction2023, xu2025cosdh}, which has emerged as the dominant strategy by striking a balance between these extremes. In this paradigm, intermediate feature representations, often the BEV feature, are transmitted, allowing for richer information sharing at a moderate communication cost. A common assumption, initiated by Where2Comm~\cite{huWhere2commCommunicationEfficientCollaborative2022}, is that foreground region features (\eg vehicles, pedestrians) provide the most informative cues while keeping transmission costs manageable.

\noindent\textbf{Foreground-Centric Sharing Strategy.} As illustrated in Fig.~\ref{fig:comparison}, Where2Comm~\cite{huWhere2commCommunicationEfficientCollaborative2022} employs a confidence generator to estimate the likelihood of object presence in each BEV grid. Connected and autonomous vehicles (CAVs) then share regions where the ego has low confidence (\ie predicted as background) but neighboring CAVs are confident (\ie predicted as foreground). CoSDH~\cite{xu2025cosdh} extends this idea with a supply–demand mechanism, using point cloud density (per pillar) as a proxy for observation reliability.  Sparsely populated pillars are treated as uncertain and thus prioritized for CAVs' support. While effective, both approaches largely disregard the informative cues embedded in background regions. CORE~\cite{wangCoreCooperativeReconstruction2023} indirectly addresses this by introducing a masked modeling strategy: it selects high-activated regions for transmission and applies a self-reconstruction objective at the receiver to recover the holistic scene. Although this implicitly captures some background context, it risks reconstructing irrelevant or noisy background content, potentially undermining perception robustness.

% In contrast to these foreground-centric strategies, we explicitly acknowledge and harness background contextual information. Rather than discarding or implicitly reconstructing it, we propose a \textit{curricular background pruning strategy}, where informative background regions are selectively mined during training to complement the foreground. A progressive pruning schedule gradually reduces reliance on shared background features, encouraging the model to encapsulate contextual cues into the foreground representation itself. Consequently, at inference, our framework relies solely on foreground features which is now enriched with background context, achieving efficient and context-aware collaborative perception.

Unlike prior methods that discard (\ie Where2Comm, CoSDH) or implicitly reconstruct (\ie CORE) background, we explicitly harness contextual cues in background via two complementary ways. First, foreground features are directly enriched with holistic scene cues. Second, background context is gradually pruned through a curriculum process, guiding the model to internalize these cues into the foreground representation and enabling more effective fusion. At inference, only enriched foreground features are shared, yielding efficient yet context-aware collaborative perception.
\section{Problem Formulation}

\begin{figure*}[t] \centering
    \includegraphics[width=0.97\textwidth]{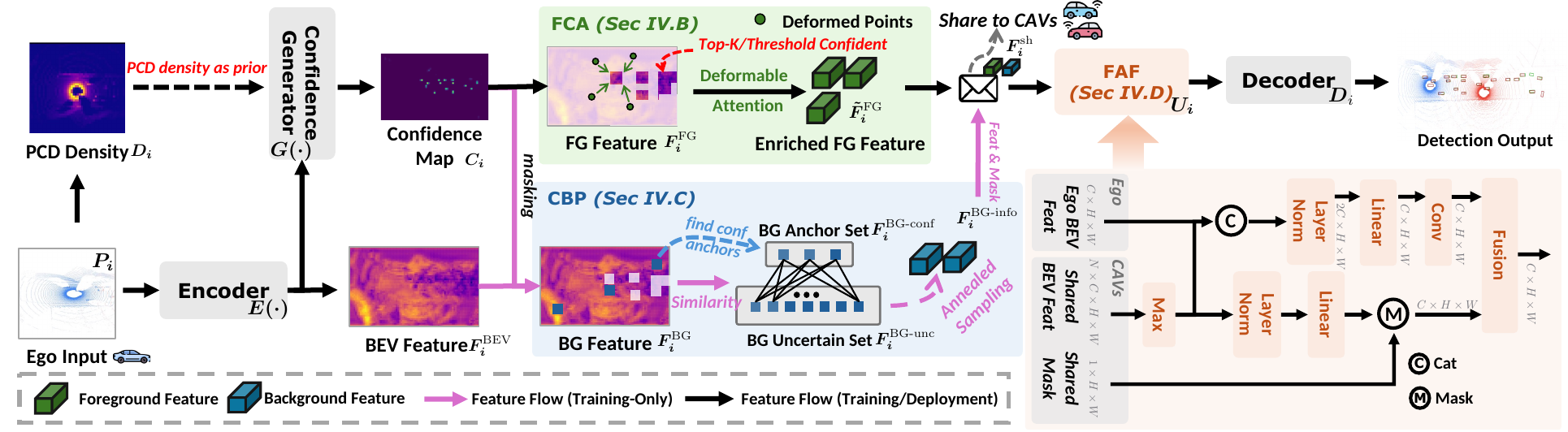}
    \caption{Overview of \framework.} \label{fig:framework}
    \vspace{-2em}
\end{figure*}

In this paper, we focus on the problem of \emph{3D object detection in collaborative perception} (CP) involving $N$ agents, each equipped with its own CP model. 
We focus on the \emph{intermediate fusion} paradigm, assuming all on-device models are trained accordingly. 
The architecture of each CP model consists of an encoder $E_i$, a compressor $\phi_i$, a decompressor $\psi_i$, a collaborative fusion module $U_i$, and a decoder $D_i$, where $i \in \{1, 2, \ldots, N\}$ indexes the agent.

Given the raw observation $X_i$, the encoder produces a BEV feature map:
\begin{equation}
    F_i = E_i(X_i), \quad F_i \in \mathbb{R}^{C \times H \times W},
\end{equation}
where $C$ is the channel dimension and $(H, W)$ the BEV spatial resolution.

Since transmitting the full BEV feature map is infeasible under vehicular bandwidth limits, the compressor applies \textit{sparse spatial sharing}:
\begin{equation}
    \tilde{F}_i = \phi_i(F_i), \quad \tilde{F}_i \subset \mathbb{R}^{C \times H \times W}.
\end{equation}
This selects a subset of grid positions in the $(H, W)$ plane and sends only their $C$-dimensional feature vectors, reducing communication overhead. The sampled feature map $\tilde{F}_i$ (extended to $F_i^{\text{sh}}$ in our method), whose communication volume (in bits) is $P_i$, is then broadcast within range $\delta$.

When agent $i$ receives $\tilde{F}_j$ from a neighbor $j$ with distance $d(i,j) \le \delta$, it recovers the dense BEV feature map via the decompressor (\eg scattering or reconstruction):
\begin{equation}
    F_j = \psi_i(\tilde{F}_j), \quad F_j \in \mathbb{R}^{C \times H \times W}.
\end{equation}
The fusion module $U_i$ then aggregates the ego feature with the recovered features from neighbors:
\begin{equation}
    F'_i = U_i\big(F_i, \{F_j \mid d(i,j) \le \delta\}\big),
\end{equation}
and the decoder outputs the final detections:
\begin{equation}
    O_i = D_i(F'_i).
\end{equation}

Given ground-truth labels $y_i$, the training objective is to maximize detection performance under a communication budget $B$:
\begin{equation}
    \xi_{\Phi}(B) = \arg\max_{\theta} \sum_{i=1}^{N} g(O_i, y_i),
\end{equation}
subject to
\begin{equation}
    \sum_{j=1}^N \big|P_{j \rightarrow i}\big| \le B,
    \label{eq:budget}
\end{equation}
where $|P_{j \rightarrow i}|$ denotes the communication volume of the message transmitted from agent $j$ and received by agent $i$, and $g(\cdot,\cdot)$ is the evaluation metric (\eg mAP).
\section{Methodology}

\subsection{Overview Framework}

As illustrated in Fig.~\ref{fig:framework}, each agent $i$ begins by encoding its local point cloud $\mathbf{P}_i$ with an encoder $E(\cdot)$ (\eg PointPillars~\cite{lang2019pointpillars}) to produce an intermediate BEV feature map $\mathbf{F}^{\text{BEV}}_i \in \mathbb{R}^{C \times H \times W}$ and a density prior $\mathbf{D}_i$. A confidence generator $G(\cdot)$ estimates a confidence map $\mathbf{C}_i \in [0,1]^{1 \times H \times W}$, which is used to partition the feature into foreground $\mathbf{F}^{\text{FG}}_i$ and background $\mathbf{F}^{\text{BG}}_i$ regions. These regions are processed by two complementary branches:  
First, the \emph{Foreground Context Attention} (FCA) branch enriches the foreground $\mathbf{F}^{\text{FG}}_i$ by refining confidence with density priors and applying deformable attention over $\mathbf{F}^{\text{BEV}}_i$, yielding contextually enhanced features $\tilde{\mathbf{F}}^{\text{FG}}_i$.  
Second, the \emph{Curricular Background Pruning} (CBP) branch (applied only during training) selects informative subsets $\mathbf{F}^{\text{BG-Info}}_i$ from the background and gradually anneals their transmission ratio. This curriculum encourages the model to encapsulate essential background context into the foreground representation.

During training, both $\tilde{\mathbf{F}}^{\text{FG}}_i$ and $\mathbf{F}^{\text{BG-Info}}_i$ are transmitted to collaborators, forming the shared feature $\mathbf{F}^{\text{sh}}_i$. At deployment, transmission is streamlined to include only the enriched foreground $\tilde{\mathbf{F}}^{\text{FG}}_i$, ensuring efficiency.  
Finally, each agent $i$ collects shared features ${\mathbf{F}^{\text{sh}}_j \mid j \in \mathcal{N}(i)}$ from its neighbors and fuses them with its ego feature via the \textit{Collaborative Context Fusion (CCF)} module. The fused representation is then decoded to produce the final detection outputs.

\subsection{Foreground Context Attention (FCA)}
\noindent\textbf{Observation 1.}  
Foreground-only sharing often suffers from false positives and incomplete semantics, as predicted foreground masks fail to account for contextual cues present in surrounding background regions.  

\noindent\textbf{Design 1.}  
To address this, FCA enhances the reliability of foreground prediction in two ways.  

\textit{First,} unlike CoSDH~\cite{xu2025cosdh}, which uses point cloud (PCD) density only to measure ego demand for communication, we incorporate point cloud density as a prior to refine the predicted confidence map: 
\[
\mathbf{C}'_i = \big( 1 - \text{norm}(\mathbf{D}_i) \big) \odot \mathbf{C}_i ,
\]
where $\mathbf{C}_i$ is the confidence map, $\mathbf{D}_i$ is the PCD density map, $\text{norm}(\cdot)$ denotes min–max normalization, and $\odot$ indicates element-wise multiplication.  
This refinement suppresses spurious false positives in low-density regions and reinforces high-density background areas as reliable negatives. 

\textit{Second,} the refined foreground features $\mathbf{F}^{\text{FG}}_i$ are further enriched via multi-scale deformable attention~\cite{zhu2020deformable} over the entire BEV map $\mathbf{F}^{\text{BEV}}_i$, producing context-aware foreground features $\tilde{\mathbf{F}}^{\text{FG}}_i$. By integrating context into the foreground, FCA mitigates semantic incompleteness and provides a more reliable foundation for subsequent feature sharing and fusion.  

\subsection{Curricular Background Pruning (CBP)}

\noindent\textbf{Observation 2.}  
While background encodes valuable context (\eg occlusion reasoning, scene layout), transmitting the entire background is infeasible. The central question is therefore how to exploit background cues during training while avoiding reliance on it at inference.

\noindent\textbf{Design 2.} 
As shown in Algorithm~\ref{alg:cbp}, CBP addresses this challenge by gradually shifting the model from background-assisted learning to foreground-only inference. At early training stages, CBP enriches learning with both foreground and informative background regions. Over time, it progressively reduces the background sharing ratio through a curriculum, forcing the model to internalize context into the shared foreground. Specifically, CBP consists of two key mechanisms:

\textbf{Step 1: Informative background mining.} 
Naively transmitting all background is wasteful, as most regions contain trivial empty space.  
To mitigate this, CBP partitions the background features $\mathbf{F}^{\text{BG}}_i$ into two disjoint subsets:  
\emph{confident background anchors} $\mathbf{F}^{\text{BG-conf}}_i$ (low-confidence, high-density areas that are almost certainly background),  
and \emph{uncertain background} $\mathbf{F}^{\text{BG-unc}}_i$ (regions where background assignment is less reliable due to sparsity, occlusion, or sensor noise).  
From the uncertain pool, CBP selects the most informative elements by measuring their similarity to confident background anchors. This procedure ensures that only the most representative background patterns are retained, so background is not discarded as noise but reinterpreted as a \emph{scaffold} grounding foreground interpretation.

\textbf{Step 2: Progressive pruning with a curriculum.}  
To avoid over-reliance on background, CBP regulates its usage through a multi-stage annealing schedule. Training begins with a higher background sharing ratio $r$, analogous to a warm start, so the model benefits from abundant contextual cues. At each stage, this ratio is decayed by a factor $\gamma$, progressively pruning background features. The gradual reduction mimics a scaffolding process: early exposure to background provides support, while later pruning encourages the model to consolidate context within its foreground representation. By the final stage, background sharing is eliminated, and the model transmits only enriched foreground features that implicitly encode contextual priors.

\begin{algorithm}[t]
\caption{Curricular Background Pruning (CBP)}
\label{alg:cbp}
\DontPrintSemicolon
\SetKwFunction{MineBG}{MineBG}
\SetKwProg{Fn}{Function}{:}{}

\KwIn{BEV features $\mathbf{F}_i$, density $\mathbf{D}_i$, confidence $\mathbf{C}_i$, 
foreground mask $\mathbf{M}^{FG}_i$, initial BG ratio $r$, similarity ratio $\tau$, decay $\gamma$, 
image size $H{\times}W$}
\KwOut{Shared feature $\mathbf{F}^{\text{sh}}_i$, transmission mask $\mathbf{M}^{\text{sh}}_i$}

\BlankLine
\textbf{Notation.} $\text{TopK}_k(\cdot|\cdot)$ selects the $k$ elements with the highest score (right argument).
$\text{sim}(\cdot,\cdot)$ denotes feature similarity (\ie cosine). \;

\BlankLine
\textcolor{teal}{\textbf{Step 1: Informative Background Mining}}\;
\Fn{\MineBG{$\mathbf{F}_i,\mathbf{D}_i,\mathbf{C}_i,\mathbf{M}^{FG}_i,r,\tau$}}{
    $\mathbf{M}^{BG}_i \leftarrow \mathbf{1} - \mathbf{M}^{FG}_i$ \tcp*{\scriptsize background mask}
    $\mathbf{C}^{BG}_i \leftarrow (1 - \mathbf{C}_i)\odot \mathbf{D}_i$ \tcp*{\scriptsize refine confidence}
    $k_a \leftarrow \lfloor r \cdot HW \rfloor$,\quad $k_\tau \leftarrow \lfloor \tau \cdot HW \rfloor$ \;
    $\mathbf{F}^{BG\text{-}conf}_i \leftarrow \text{TopK}_{k_a}(\mathbf{F}^{BG}_i \,|\, \mathbf{C}^{BG}_i)$ \tcp*{\scriptsize anchors}
    $\mathbf{F}^{BG\text{-}unc}_i \leftarrow \mathbf{F}^{BG}_i \setminus \mathbf{F}^{BG\text{-}conf}_i$ \tcp*{\scriptsize uncertain pool}
    $\mathbf{F}^{BG\text{-}sel}_i \leftarrow \text{TopK}_{k_\tau}\!\big(\text{sim}(\mathbf{F}^{BG\text{-}unc}_i,\mathbf{F}^{BG\text{-}conf}_i)\big)$ \;
    $\mathbf{M}^{\text{sh}}_i \leftarrow \mathbf{M}^{FG}_i \cup \mathbf{M}^{BG\text{-}sel}_i$ \tcp*{\scriptsize FG + mined BG}
    \KwRet $\mathbf{M}^{\text{sh}}_i$
}

\BlankLine
\textcolor{teal}{\textbf{Step 2: Progressive Pruning (Curriculum)}}\;
\For{epoch $=1,\dots,E$}{
    \For{each training step / minibatch}{
        $\mathbf{M}^{\text{sh}}_i \leftarrow \MineBG(\mathbf{F}_i,\mathbf{D}_i,\mathbf{C}_i,\mathbf{M}^{FG}_i,r,\tau)$ \;
        $\mathbf{F}^{\text{sh}}_i \leftarrow \mathbf{F}_i \odot \mathbf{M}^{\text{sh}}_i$
    }
    
    $r \leftarrow \gamma \cdot r$ \tcp*{\scriptsize anneal BG ratio }
}
\KwRet $\mathbf{F}^{\text{sh}}_i, \mathbf{M}^{\text{sh}}_i$
\end{algorithm}

\subsection{Foreground Amplification Fusion (FAF)}

\noindent\textbf{Observation 3.}  
Collaboration improves perception only if shared cues are fused effectively with the ego representation. Simple strategies such as concatenation or averaging are problematic: irrelevant or unaligned regions from collaborators introduce noise, and distributional mismatch between ego and remote features blurs salient activations. Without careful design, fusion can degrade the feature map, making it less distinguishable rather than more informative.

\noindent\textbf{Design 3.}  
FAF addresses this by enforcing two principles: (1) fusion should explicitly model interactions between ego and received features, and (2) collaborative updates should amplify foreground cues while suppressing background noise.

Formally, let the ego feature be $\mathbf{F}^{\text{ego}}_i \in \mathbb{R}^{C \times H \times W}$, and let each neighbor $j \in \mathcal{N}(i)$ broadcast a shared feature $\mathbf{F}^{\text{sh}}_j$ with mask $\mathbf{M}^{\text{sh}}_j$. At each timestep, ego $i$ aggregates the received features by element-wise max fusion:
\[
\mathbf{F}^r_i = \text{Proj}\Big(\text{LN}(\max_{j \in \mathcal{N}(i)} \mathbf{F}^{\text{sh}}_j)\Big), \quad
\mathbf{M}^{\text{sh}}_i = \max_{j \in \mathcal{N}(i)} \mathbf{M}^{\text{sh}}_j .
\]

\textbf{Interaction Modeling.} 
The normalized neighbor feature $\mathbf{F}^r_i$ is combined with the ego representation to model cross-agent interactions:
\[
\mathbf{F}^{\text{merge}}_i 
= \text{Conv}\!\Big(\text{Proj}\big(\text{LN}([\mathbf{F}^{\text{ego}}_i, \mathbf{F}^r_i])\big)\Big).
\]
Here, layer normalization equalizes activation scales, ensuring that foreground regions are not drowned out by noisy background responses.

\textbf{Foreground Amplification.} 
Finally, the merged feature is gated by the transmission mask and residually added to the ego feature:
\[
\mathbf{F}^{\text{fused}}_i 
= \mathbf{F}^{\text{ego}}_i 
+ \text{Proj}\!\Big(\mathbf{F}^{\text{merge}}_i 
\odot \mathbf{F}^r_i 
\odot \mathbf{M}^{\text{sh}}_i\Big).
\]
This step selectively reinforces salient foreground activations while suppressing spurious background signals, ensuring that collaboration strengthens the representation rather than injecting noise. The result is a cleaner, more discriminative fused feature (visualized in Fig.~\ref{fig:topk_visualize} later).

\section{Evaluation}

\begin{table*}[t]
\centering
\caption{Detection accuracy and information selection ratio on \textcolor{red!70}{OPV2V}~\cite{opv2v}, \textcolor{red!70}{V2X-R}~\cite{V2X-R}, and \textcolor{blue!70}{DAIR-V2X}~\cite{dair-v2x} datasets. ``Ratio'' denotes the proportion of BEV-plane information selected per collaborative agent. \textbf{Values} represent the best performance across all settings. \textcolor{teal}{Values} indicate the performance gain when increasing the ratio from 1\% to 10\%, while \textcolor{red}{values} denote the improvement over the second-best method at the same ratio. $^*$ We adopt the intermediate fusion variant of CoSDH~\cite{xu2025cosdh}.}
\resizebox{\textwidth}{!}{
\renewcommand{\arraystretch}{1.5}
\begin{tabular}{cc|cc|cc|cc}
\hline
\multicolumn{2}{c|}{\multirow{2}{*}{Method}} & \multicolumn{2}{c|}{\textcolor{red!70}{OPV2V}~\cite{opv2v}} & \multicolumn{2}{c|}{\textcolor{red!70}{V2X-R}~\cite{V2X-R}} & \multicolumn{2}{c}{\textcolor{blue!70}{DAIR-V2X}~\cite{dair-v2x}} \\
\cline{3-8}
& & AP@0.3/0.5/0.7$\uparrow$ & Ratio$\downarrow$ & AP@0.3/0.5/0.7$\uparrow$ & Ratio$\downarrow$ & AP@0.3/0.5/0.7$\uparrow$ & Ratio$\downarrow$  \\
\hline
\multicolumn{8}{c}{\cellcolor{gray!10}\textbf{Basic}} \\
\hline
% NO FUSION
~\textcolor[HTML]{708090}{\scriptsize{[\phantom{xxxx}-\phantom{xxxx}]}} & 
No Fusion & 84.95/83.72/73.93 & 0\% & 72.52/71.09/61.06 & 0\%  &  70.26/67.05/57.05 & 0\%  \\
% EARLY FUSION
~\textcolor[HTML]{708090}{\scriptsize{[\phantom{xxxx}-\phantom{xxxx}]}} & 
Early Fusion & 94.95/94.52/87.81 & 100\% & 88.17/87.82/77.73 & 100\% & 76.53/71.93/56.77 & 100\% \\
% LATE FUSION
~\textcolor[HTML]{708090}{\scriptsize{[\phantom{xxxx}-\phantom{xxxx}]}} & 
Late Fusion & 95.62/94.62/88.76 & - & 86.58/85.79/78.22 & - & 78.07/65.60/47.53 & -  \\
\hline
\multicolumn{8}{c}{\cellcolor{gray!10}\textbf{Dense Spatial Sharing}} \\
\hline
% When2Comm
~\textcolor[HTML]{708090}{\scriptsize{[CVPR 2020]}} & 
When2Comm~\cite{liuWhen2comMultiAgentPerception2020} & 
89.12/87.14/67.28 & 100\% & 
86.01/84.74/74.75 & 100\% & 
66.31/58.45/36.74 & 100\% \\
% V2VNet
~\textcolor[HTML]{708090}{\scriptsize{[ECCV 2020]}} & 
V2VNet~\cite{wangV2VNetVehicletoVehicleCommunication2020} & 94.13/92.94/79.79 & 100\% & 85.12/83.65/67.92 & 100\% & 78.28/74.04/56.14 & 100\% \\
% V2XViT
~\textcolor[HTML]{708090}{\scriptsize{[ECCV 2022]}} & 
V2XViT~\cite{xuV2XViTVehicletoEverythingCooperative2022} & 94.89/91.06/68.25 & 100\% & 91.17/88.93/78.19 & 100\% & 83.03/77.54/60.84 & 100\%  \\

% AttnFuse
~\textcolor[HTML]{708090}{\scriptsize{[ICRA 2022]}} &
AttFuse~\cite{opv2v} & \textbf{96.41}/94.79/81.05 & 100\% & 85.86/81.30/54.82 & 100\% & 82.99/78.72/62.02 & 100\% \\

\hline
\multicolumn{8}{c}{\cellcolor{gray!10}\textbf{Sparse Spatial Sharing}} \\
\hline
% Where2Comm
\multirow{4}{*}{\textcolor[HTML]{708090}{\scriptsize [NIPS 2022]}} &
\multirow{4}{*}{Where2Comm~\cite{huWhere2commCommunicationEfficientCollaborative2022}} 
% row1
& 93.45/92.70/84.39 & 1\%
& 90.24/89.09/77.55 & 1\%
& 80.83/76.35/62.01 & 1\% \\
% row2
& & 94.83/94.19/86.26& 5\% 
& 91.74/90.47/79.37 & 5\% 
& 82.10/77.67/63.09 & 5\% \\
% row3
& & 94.97/94.32/86.50 & 10\% 
& 92.00/90.68/79.76 & 10\%  
& 82.19/77.81/63.19 & 10\% \\
% row4
& & \cellcolor{green!5}\textcolor{teal}{+1.52/+1.62/+2.11} & \scriptsize\cellcolor{green!5}1\% $\rightarrow$ 10\% &
\cellcolor{green!5}\textcolor{teal}{+1.76/+1.59/+2.21} & \scriptsize\cellcolor{green!5} 1\% $\rightarrow$ 10\% &
\cellcolor{green!5}\textcolor{teal}{+1.36/+1.46/+1.18} & \scriptsize\cellcolor{green!5} 1\% $\rightarrow$ 10\% \\
\hdashline

% CORE
\multirow{4}{*}{\textcolor[HTML]{708090}{\scriptsize [ICCV 2023]}} &
\multirow{4}{*}{CORE~\cite{wangCoreCooperativeReconstruction2023}} &
% row 1
51.52/50.46/40.72& 1\% &
39.21/38.24/30.61 & 1\% &
44.48/42.27/32.59 & 1\%\\
% row2
& & 80.96/79.59/67.18 & 5\% &
75.53/74.14/62.98 & 5\% &
59.83/56.16/44.60 & 5\% \\
% row 3
& & 86.10/84.57/71.19 & 10\% &
82.09/80.42/68.53 & 10\% &
64.26/60.09/47.59 & 10\% \\
% row4
& & \cellcolor{green!5}\textcolor{teal}{+34.58/+34.11/+30.47} & \scriptsize\cellcolor{green!5}1\% $\rightarrow$ 10\% &
\cellcolor{green!5}\textcolor{teal}{+42.88/+42.18/+37.92} & \scriptsize\cellcolor{green!5} 1\% $\rightarrow$ 10\% &
\cellcolor{green!5}\textcolor{teal}{+19.78/+17.82/+15.00} & \scriptsize\cellcolor{green!5} 1\% $\rightarrow$ 10\% \\
\hdashline

% CoSDH
\multirow{4}{*}{\textcolor[HTML]{708090}{\scriptsize [CVPR 2025]}} &
\multirow{4}{*}{CoSDH$^*$~\cite{xu2025cosdh}} &
% row1
90.54/89.06/77.70 & 1\% &
84.78/83.38/74.19 & 1\% &
78.89/74.40/61.30 & 1\% \\
% row2
& & 92.40/90.99/79.61 & 5\% &
84.74/83.33/74.12 & 5\% &
82.10/77.89/64.02 & 5\%  \\
% row3
& & 93.89/92.74/81.67 & 10\% &
84.77/83.34/74.09 & 10\% &
82.72/78.57/64.35 & 10\% \\
% row4
& & \cellcolor{green!5}\textcolor{teal}{+3.35/+3.68/+3.97} & \scriptsize\cellcolor{green!5}1\% $\rightarrow$ 10\% &
\cellcolor{green!5}\textcolor{teal}{-0.01/-0.04/-0.10} & \scriptsize\cellcolor{green!5} 1\% $\rightarrow$ 10\% &
\cellcolor{green!5}\textcolor{teal}{+3.83/+4.17/+3.05} & \scriptsize\cellcolor{green!5} 1\% $\rightarrow$ 10\% \\

% OURS
\hdashline
\multirow{4}{*}{\textcolor[HTML]{708090}{\scriptsize [\phantom{xxx}OURS\phantom{xxx}]}} &
\multirow{4}{*}{\framework} &
% row1
\rule{0pt}{4ex}\makecell{95.81/95.05/88.10 \\ \scriptsize\textcolor{red}{(+2.36/+2.35/+3.71)}} & 1\% &
\makecell{92.21/90.88/81.23 \\ \scriptsize\textcolor{red}{(+1.97/+1.79/+4.68)}} & 1\% &
\makecell{83.27/79.10/64.45 \\ \scriptsize\textcolor{red}{(+2.44/+2.75/+2.44)}} & 1\% \\
% row2
& & \makecell{96.01/95.39/88.99 \\ \scriptsize\textcolor{red}{(+1.18/+1.07/+2.49)}} & 5\% &
\makecell{92.90/91.71/83.25 \\ \scriptsize\textcolor{red}{(+1.16/+1.24/+3.88)}} & 5\% &
\makecell{\textbf{83.65}/\textbf{79.87}/65.94 \\ \scriptsize\textcolor{red}{(+1.55/+1.98/+1.92)}} & 5\% \\[1ex]
% row3
& & \makecell{96.01/\textbf{95.41}/\textbf{89.02} \\ \scriptsize\textcolor{red}{(+1.04/+1.09/+2.52)}} & 10\% &
\makecell{\textbf{92.97}/\textbf{91.79}/\textbf{83.32} \\ \scriptsize\textcolor{red}{(+0.97/+1.11/+3.56)}} & 10\% &
\makecell{83.52/79.76/\textbf{65.98} \\ \scriptsize\textcolor{red}{(+0.8/+1.19/+1.63)}} & 10\% \\[1ex]
% row4
& & \cellcolor{green!5}\textcolor{teal}{+0.20/+0.36/+0.92} & \scriptsize\cellcolor{green!5}1\% $\rightarrow$ 10\% &
\cellcolor{green!5}\textcolor{teal}{+0.06/+0.91/+2.09} & \scriptsize\cellcolor{green!5} 1\% $\rightarrow$ 10\% &
\cellcolor{green!5}\textcolor{teal}{+0.25/+0.66/+1.53} & \scriptsize\cellcolor{green!5} 1\% $\rightarrow$ 10\%\\

\hline
\end{tabular}
}
\begin{flushleft}        
    \scriptsize Bandwidth consumption (100\% for dense, 1/5/10\% for sparse): V2VNet = V2XViT = AttFuse $>$ When2Comm $>$ Early $>$ realworld bandwidth constraint 28Mbps~\cite{dsrc} $>$ Where2Comm = CORE $>$ CoSDH =  \framework $>$ Late. Though bandwidth-efficient, late fusion inevitably incurs additional latency (\eg full model inference, post-processing).
\end{flushleft}
\vspace{-2.5em}
\label{tab:result}
\end{table*}

\subsection{Experimental Setup}

\noindent\textbf{Datasets.} We evaluate our method on both \textcolor{red!70}{simulated} and \textcolor{blue!70}{real-world} datasets (\textcolor{red!70}{OPV2V}~\cite{opv2v}, \textcolor{red!70}{V2X-R}~\cite{V2X-R} and \textcolor{blue!70}{DAIR-V2X}~\cite{dair-v2x}) for LiDAR-based 3D object detection. These datasets span varying numbers of collaborative agents, object density, and lane conditions.

\noindent\textbf{Baselines.} We compare our method against recent open-source approaches, including (1) naive settings, (2) dense spatial sharing methods that transmit the full intermediate BEV feature (\ie When2Comm~\cite{liuWhen2comMultiAgentPerception2020}, V2VNet, V2XViT~\cite{xuV2XViTVehicletoEverythingCooperative2022}, and AttFuse~\cite{opv2v}) and (3) sparse sharing methods that follow a foreground-centric strategy (\ie Where2Comm~\cite{huWhere2commCommunicationEfficientCollaborative2022}, CORE~\cite{wangCoreCooperativeReconstruction2023}, and CoSDH~\cite{xu2025cosdh}). For CoSDH, which employs an intermediate–late hybrid fusion strategy, we focus on the intermediate fusion variant, as late fusion is less practical for real-world deployment (especially in high-speed scenarios) due to the additional latency introduced by complete model inference and post-processing (\eg NMS).

\begin{figure*}[t] \centering
    \includegraphics[width=\textwidth]{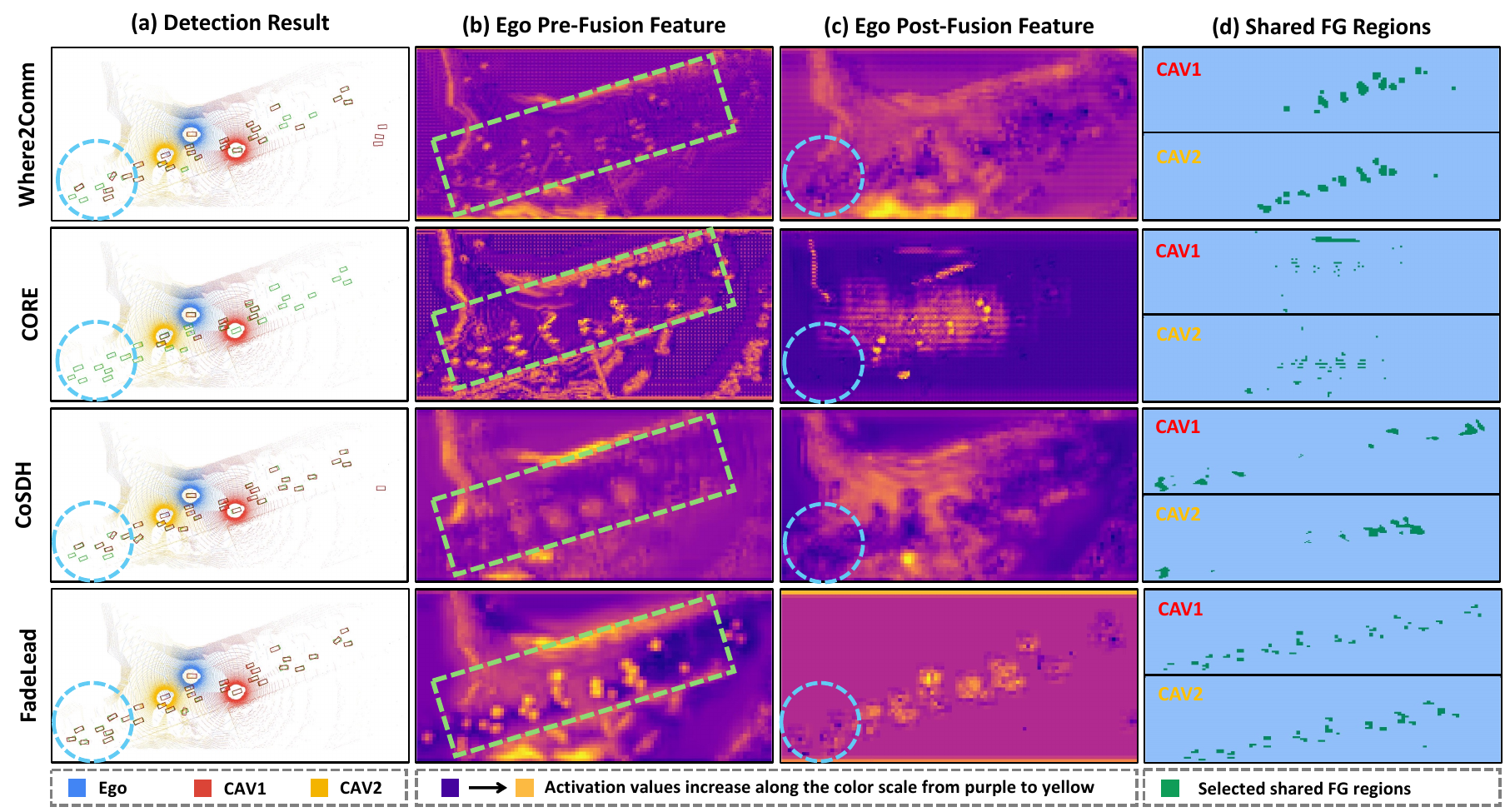}
    \caption{Visualization on OPV2V~\cite{opv2v} with top-1\% confident foreground selection. We show (a) detection results, (b) the ego BEV feature before fusion, (c) the ego BEV feature after fusion, and (d) the shared regions from collaborating CAVs.} \label{fig:qualitative}
    \vspace{-2em}
\end{figure*}

\begin{figure}[t] \centering
    \includegraphics[width=0.48\textwidth]{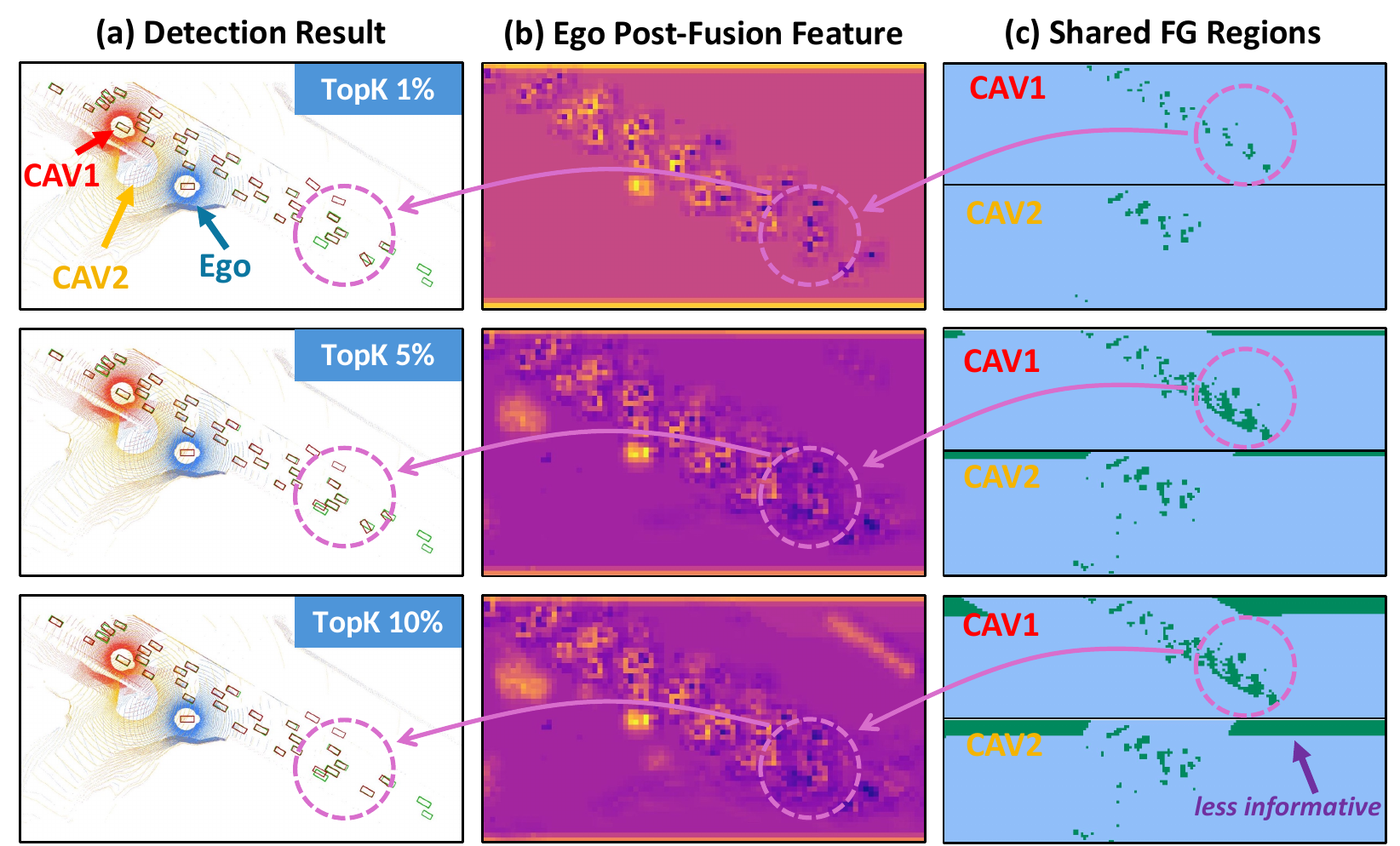}
    \caption{Visualization of detection results, post-fusion features, and shared FG regions under different top-$k$ ratios on V2X-R~\cite{V2X-R}.}
    \label{fig:topk_visualize}

\end{figure}

\noindent\textbf{Implementation.} For fair comparison, we standardize the spatial coverage and resolution of intermediate BEV features\footnote{This standardization may result in slight performance differences compared to the reported results in the original paper.}. Specifically, OPV2V and V2X-R adopt a perception range of [-140.8, -38.4, -3, 140.8, 38.4, 1] with a downsampling rate of 4, while DAIR-V2X uses [-100.8, -40, -3.5, 100.8, 40, 1.5] with a downsampling rate of 2. All baseline methods are retrained with their original hyperparameters and use PointPillars~\cite{lang2019pointpillars} with voxel size [0.4, 0.4, h]m as the encoder. Under these settings, the resulting BEV feature map resolutions are $176\times48$ for OPV2V/V2X-R and $126\times50$ for DAIR-V2X. For CoSDH and \framework, we apply channel compression that downsamples the shared feature map from \([H, W, C]\) to \([H, W, C/R]\) (C=256 and R=16 in our settings) along with quantization (float32$\rightarrow$float16), which further reduces bandwidth consumption. To better reflect real-world deployment constraints, we avoid projecting point clouds into the ego coordinate during preprocessing (a step adopted in some prior works for training stability, but impractical before feature sharing in real deployments). For CBP, we set ($r$=0.1, $\gamma$=0.8) with decay every 5 epochs. Full hyperparameters and configurations are provided in anonymously released code. All models are trained on four Nvidia RTX 3090 GPUs.

\noindent\textbf{Evaluation.} We evaluate average precision (AP) at IoU thresholds of 0.3, 0.5 and 0.7 for 3D object detection. To examine the effectiveness of spatial information selection, we vary the selection ratio to 1\%, 5\%, and 10\%. These values are chosen as the foreground typically occupies  $\leq10\%$ of the BEV plane across all scenarios, and the resulting bandwidth remains within the limits of vehicular networks~\cite{dsrc}.

\vspace{-0.5em}
\subsection{Quantitative Results}

\noindent\textbf{Better performance even under extreme low network bandwidth.} 
We evaluate \framework against baseline methods using AP@0.3/0.5/0.7 under varying information selection ratios. As shown in Table~\ref{tab:result}, dense spatial sharing methods (\ie When2Comm~\cite{liuWhen2comMultiAgentPerception2020}, V2VNet~\cite{wangV2VNetVehicletoVehicleCommunication2020}, V2XViT~\cite{xuV2XViTVehicletoEverythingCooperative2022}, AttFuse~\cite{opv2v}) can achieve strong detection accuracy but only by transmitting the entire BEV feature, making them bandwidth-inefficient for practical deployment. In contrast, sparse sharing approaches transmit only a small fraction of informative regions while often achieving comparable or superior accuracy at significantly reduced bandwidth cost.
Within the sparse sharing category, \framework consistently outperforms SOTA sparse spatial sharing methods across all datasets at corresponding selection ratios, as indicated by the \textcolor{red}{values} in Table~\ref{tab:result}. At the most restrictive 1\% selection ratio, \framework demonstrates substantial improvements: +2.36/+2.35/+3.71 AP on OPV2V, +1.97/+1.79/+4.68 AP on V2X-R, and +2.44/+2.75/+2.44 AP on DAIR-V2X compared to the second-best method. This low bandwidth requirement is particularly advantageous in dense collaborative scenarios (\eg intersections with multiple participating vehicles) and provides better resilience in unstable vehicular networks. 

\noindent\textbf{Background contextual cues are effectively encapsulated.}
The \textcolor{teal}{values} in Table~\ref{tab:result} show that \framework reaches near-optimal performance even at the  1\% selection ratio, with only marginal improvements when increasing the ratio to 10\% (\eg +0.20/+0.36/+0.92 on OPV2V). This small gap indicates that essential background context is already encapsulated into the minimal set of shared foreground regions, leaving little room for further gains by transmitting more background. In contrast, reconstruction-based method CORE exhibits severe instability, with large fluctuations across ratios (\eg +34.58/+34.11/+30.47 on OPV2V). The stability of \framework demonstrates that it integrates contextual cues efficiently at extremely low bandwidth, providing both robustness and scalability for real-world deployment.

\noindent\textbf{\framework with improved bandwidth.} \framework is primarily designed for extreme bandwidth constraints, where only predicted foreground region features are transmitted. A natural question, however, is whether additional bandwidth can further enhance its performance. 
% As shown in Fig.~\ref{fig:topk_visualize}, increasing the top-$k$ ratio to first tends to include a small portion of background features adjacent to the predicted foreground and later more it tends to include more regions in the boarder (more centain BG regions) provides more complementary context, which can help correct feature maps and refine detection results.
As shown in Fig.~\ref{fig:topk_visualize}c, increasing the top-$k$ ratio first brings in small portions of background adjacent to the foreground (\ie 1\%$\rightarrow$5\%), which provide useful complementary context to correct feature maps. As the ratio grows further (\ie 5\%$\rightarrow$10\%), more certain background regions (typically near scene boundaries) are included. While less informative, these regions can still offer minor benefits for refining detection results.

\begin{figure}[t] \centering
    \includegraphics[width=0.57\textwidth]{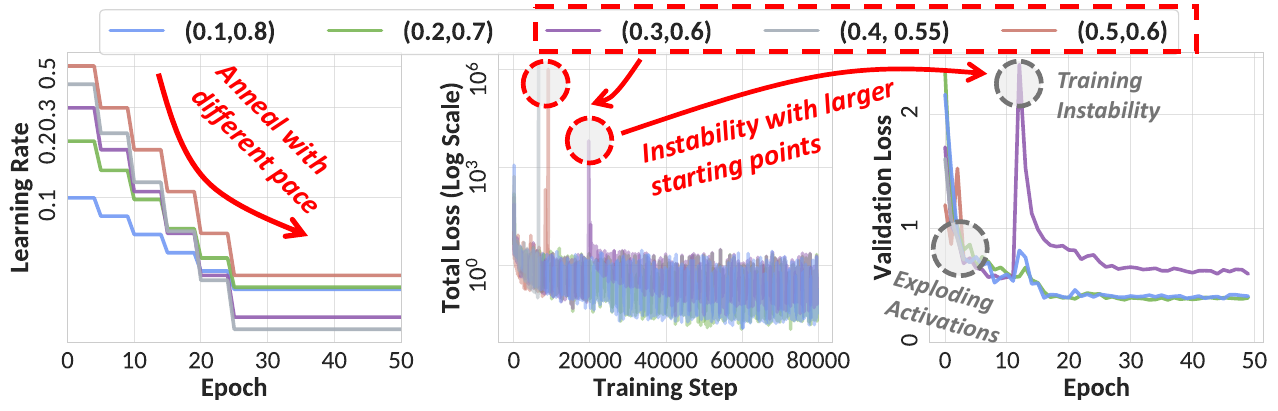}
    \caption{Effect of curriculum pacing under different initialization points and progression rates on OPV2V~\cite{opv2v}.} \label{fig:trainig_stats}
\end{figure}

\noindent\textbf{Pace sensitivity of curriculum.}  
We evaluate the sensitivity of the curriculum schedule by varying the $\big($initial background ratio $r$, the decay factor $\gamma$$\big)$. All schedules are designed to decay at comparable rates and eventually converge to a near-zero background ratio. As illustrated in Fig.~\ref{fig:trainig_stats}, larger initial ratios induce training instabilities (\eg sudden spikes in validation loss, numerical divergence), whereas smaller ratios lead to stable convergence. These results suggest that introducing a modest amount of background early on provides useful context to stabilize learning, but excessive background may overshadow foreground and disrupt training.

% \noindent\textbf{Pace sensitivity of curriculum.}  
% We evaluate the sensitivity of the curriculum schedule by varying the initial background ratio $r$ and the decay factor $\gamma$. All schedules decay at comparable rates and converge to a near-zero background ratio. As shown in Fig.~\ref{fig:trainig_stats}, larger initial ratios often cause training instabilities (\eg spikes in validation loss, numerical divergence), while smaller ratios yield stable convergence. These results indicate that introducing a modest amount of background early on can stabilize learning, but excessive background introduces noise that disrupts training.

\subsection{Qualitative Results}

\noindent\textbf{\framework enhances foreground representations.} Figure~\ref{fig:qualitative}b \rectvis{} shows that our FCA module strengthens the semantic quality of the ego’s pre-fusion features. Compared with Where2Comm~\cite{huWhere2commCommunicationEfficientCollaborative2022} and CoSDH~\cite{xu2025cosdh}, \framework produces sharper activations at target object locations, reflecting better semantic alignment. Unlike CORE~\cite{wangCoreCooperativeReconstruction2023}, which suppresses background responses uniformly, \framework retains higher activation in nearby background regions where the ego has strong observability. Though these are background areas, maintaining confident activation is beneficial: it encodes the ego’s reliable observations and helps distinguish certain background from uncertain or occluded regions. This not only improves pre-fusion representations but also provides a more trustworthy input to the subsequent collaborative fusion stage, ensuring that the shared and fused features are grounded in clear observability semantics.

\noindent\textbf{\framework enables more reliable feature sharing and fusion.}
Figure~\ref{fig:qualitative} \circledvis{} highlights how different collaborative perception methods vary in foreground prediction accuracy, feature-map quality after fusion, and the reliability of shared regions. For foreground estimation and sharing, as shown in Fig.~\ref{fig:qualitative}d, Where2Comm often misses distant objects due to inaccurate foreground prediction, leading to incomplete sharing and weak fusion. CORE’s reconstruction-based design relies heavily on abundant source information, and under limited input, it can even corrupt the ego’s pre-fusion feature map. CoSDH leverages PCD density as demanding, but many of its selected regions remain ambiguous to the ego, resulting in weak post-fusion activations. In contrast, \framework combines FCA’s explicit supervision with PCD density–smoothed confidence to produce more reliable region selection, and with FAF reinforcing these cues after fusion, its post-fusion feature map (Fig.~\ref{fig:qualitative}c) becomes more clearly distinguishable between background and foreground.

\begin{table}[t]\centering
    \caption{Ablation study.} 
    \label{tab:ablation_study}
    \hspace{3.5em}
    \resizebox{0.42\textwidth}{!}{
    \Large
    \begin{tabular}{*{3}{c}||*{3}{c}||*{3}{c}}
        \toprule
        \multicolumn{3}{c||}{} & \multicolumn{3}{c||}{OPV2V~\cite{opv2v}} & \multicolumn{3}{c}{DAIR-V2X~\cite{dair-v2x}}  \\
        \textbf{FCA} & \textbf{CBP} & \textbf{FAF}  & AP@0.3 & AP@0.5 & AP@0.7 & AP@0.3 & AP@0.5 & AP@0.7 \\
        \midrule
         $\times$/\checkmark  & $\times$  &  $\times$  & collapse & collapse & collapse &  collapse & collapse & collapse \\
        \checkmark & $\times$   & \checkmark  & collapse & collapse & collapse & \makecell{80.45 \\ \textcolor{gray}{(-3.07)}} & \makecell{76.15 \\ \textcolor{gray}{(-3.61)}} & \makecell{61.84 \\ \textcolor{gray}{(-4.14)}} \\
        $\times$   & \checkmark & \checkmark & \makecell{96.22 \\ \textcolor{gray}{(-0.21)}} & \makecell{95.17 \\ \textcolor{gray}{(-0.24)}} & \makecell{87.12 \\ \textcolor{gray}{(-1.9)}}& \makecell{83.24 \\ \textcolor{gray}{(-0.28)}} & \makecell{79.19 \\ \textcolor{gray}{(-0.57)}} & \makecell{64.91 \\ \textcolor{gray}{(-1.07)}} \\ 
        \checkmark & \checkmark & $\times$  & \makecell{\textbf{96.30} \\ \textcolor{gray}{(+0.29)} } & \makecell{94.19 \\ \textcolor{gray}{(-1.22)} } & \makecell{84.13 \\ \textcolor{gray}{(-4.89)} } & \makecell{79.82 \\ \textcolor{gray}{(-3.7)}} & \makecell{73.54 \\ \textcolor{gray}{(-6.22)}} & \makecell{57.83 \\ \textcolor{gray}{(-8.15)}} \\
        \rowcolor{gray!20}
        \checkmark  &\checkmark   & \checkmark   & 96.01 & \textbf{95.41} & \textbf{89.02} & \textbf{83.52} & \textbf{79.76} & \textbf{65.98} \\ 
        \bottomrule
    \end{tabular}
    }
\vspace{-0.5em}
\end{table}

\section{Ablation \& Discussion}
\label{sec:discussion}

\noindent\textbf{Ablation study.}
Table~\ref{tab:ablation_study} validates the effect of each component in \framework. Without CBP, strictly sharing predicted foreground features is prone to collapse during early training. Even when convergence is achieved, excluding CBP causes notable performance drop on DAIR-V2X, as the model loses the gradual curriculum that both regularizes learning and injects background context into the foreground representation. Removing FCA degrade performance especially at higher IoUs, indicating that FCA enhances precise localization by aggregating contextual cues around foreground objects (see Fig.~\ref{fig:topk_visualize}). Disabling FAF yields the sharpest decrease, particularly on DAIR-V2X, underscoring its role in effectively fusing local and shared features.

\noindent\textbf{Hybrid with late fusion.} While intermediate–late hybrid fusion inevitably incurs additional latency (\eg full model inference, post-processing) and is therefore less practical for real-world deployment, Table~\ref{tab:intermediate-late} shows that our method achieves comparable performance to the state-of-the-art CoSDH~\cite{xu2025cosdh} at AP@0.3 and AP@0.5, and consistently outperforms it at higher IoU thresholds (\ie AP@0.7).

\noindent\textbf{Further results.} Due to limited space, additional experimental details at top-5\%/10\% will be provided as supplement.

\begin{table}[t]\centering
    \caption{Performance of intermediate-late fusion variant.}
    \label{tab:intermediate-late}
    \resizebox{0.48\textwidth}{!}{
    \Large
    \begin{tabular}{c||*{3}{c}||*{3}{c}||*{3}{c}}
        \toprule
        & \multicolumn{3}{c||}{\textbf{OPV2V}} & \multicolumn{3}{c||}{\textbf{V2X-R}} & \multicolumn{3}{c}{\textbf{DAIR-V2X}} \\
        Method  & AP@0.3 & AP@0.5 & AP@0.7 & AP@0.3 & AP@0.5 & AP@0.7 & AP@0.3 & AP@0.5 & AP@0.7 \\
        \midrule
        Where2Comm & 96.43 & 96.01 & 90.40 & 93.60 & 92.55 & 83.08 & 83.26 & 78.29 & 64.03  \\
        CORE & 95.41 & 94.67 & 87.62 & 90.81 & 89.64 & 80.73 & 62.14 & 58.32 & 46.82 \\
        CoSDH & 97.36 & 96.75 & 90.79 & 92.39 & 91.72 & 85.03 & 84.59 & 79.99 & 66.01  \\
        \framework & 96.82 & 96.48 & \textbf{92.51} & 93.56 & 92.77 & \textbf{86.89} & 83.98 & 79.66 & \textbf{66.51}  \\
        \bottomrule
    \end{tabular}
    }
\end{table}

\section{Conclusion}

We presented \framework, a foreground-centric collaborative perception framework that enriches transmitted foreground region features with context to enable efficient and robust collaborative perception. Through a curricular training strategy that internalizes background context into foreground representations and fusion mechanisms that emphasize salient information, \framework push the limits of foreground-centric sharing. Extensive experiments on both simulation and real-world datasets validate its effectiveness under stringent communication constraints, underscoring its efficiency for deployment in practical vehicular networks.

\bibliographystyle{IEEEtran}
\bibliography{reference}

@misc{chenCooperCooperativePerception2019,
  title = {Cooper: {{Cooperative Perception}} for {{Connected Autonomous Vehicles}} Based on {{3D Point Clouds}}},
  shorttitle = {Cooper},
  author = {Chen, Qi and Tang, Sihai and Yang, Qing and Fu, Song},
  year = {2019},
  month = may,
  number = {arXiv:1905.05265},
  eprint = {1905.05265},
  primaryclass = {cs},
  publisher = {arXiv},
  doi = {10.48550/arXiv.1905.05265},
  urldate = {2024-12-18},
  archiveprefix = {arXiv},
  langid = {english}
}

@inproceedings{gao2025langcoop,
  title={Langcoop: Collaborative driving with language},
  author={Gao, Xiangbo and Wu, Yuheng and Wang, Rujia and Liu, Chenxi and Zhou, Yang and Tu, Zhengzhong},
  booktitle={Proceedings of the Computer Vision and Pattern Recognition Conference},
  pages={4226--4237},
  year={2025}
}

@misc{gaoSTAMPScalableTask2025,
  title = {{{STAMP}}: {{Scalable Task And Model-agnostic Collaborative Perception}}},
  shorttitle = {{{STAMP}}},
  author = {Gao, Xiangbo and Xu, Runsheng and Li, Jiachen and Wang, Ziran and Fan, Zhiwen and Tu, Zhengzhong},
  year = {2025},
  month = jan,
  number = {arXiv:2501.18616},
  eprint = {2501.18616},
  primaryclass = {cs},
  publisher = {arXiv},
  doi = {10.48550/arXiv.2501.18616},
  urldate = {2025-02-23},
  archiveprefix = {arXiv},
  langid = {english}
}

@misc{huangActFormerScalableCollaborative2024,
  title = {{{ActFormer}}: {{Scalable Collaborative Perception}} via {{Active Queries}}},
  shorttitle = {{{ActFormer}}},
  author = {Huang, Suozhi and Zhang, Juexiao and Li, Yiming and Feng, Chen},
  year = {2024},
  month = mar,
  number = {arXiv:2403.04968},
  eprint = {2403.04968},
  primaryclass = {cs},
  publisher = {arXiv},
  doi = {10.48550/arXiv.2403.04968},
  urldate = {2024-12-05},
  archiveprefix = {arXiv},
  langid = {english}
}

@misc{huWhere2commCommunicationEfficientCollaborative2022,
  title = {Where2comm: {{Communication-Efficient Collaborative Perception}} via {{Spatial Confidence Maps}}},
  shorttitle = {Where2comm},
  author = {Hu, Yue and Fang, Shaoheng and Lei, Zixing and Zhong, Yiqi and Chen, Siheng},
  year = {2022},
  month = sep,
  number = {arXiv:2209.12836},
  eprint = {2209.12836},
  primaryclass = {cs},
  publisher = {arXiv},
  doi = {10.48550/arXiv.2209.12836},
  urldate = {2024-12-09},
  archiveprefix = {arXiv},
  langid = {english}
}

@misc{liuWhen2comMultiAgentPerception2020,
  title = {When2com: {{Multi-Agent Perception}} via {{Communication Graph Grouping}}},
  shorttitle = {When2com},
  author = {Liu, Yen-Cheng and Tian, Junjiao and Glaser, Nathaniel and Kira, Zsolt},
  year = {2020},
  month = jun,
  number = {arXiv:2006.00176},
  eprint = {2006.00176},
  primaryclass = {cs},
  publisher = {arXiv},
  doi = {10.48550/arXiv.2006.00176},
  urldate = {2024-12-09},
  archiveprefix = {arXiv},
  langid = {english}
}

@misc{luExtensibleFrameworkOpen2024,
  title = {An {{Extensible Framework}} for {{Open Heterogeneous Collaborative Perception}}},
  author = {Lu, Yifan and Hu, Yue and Zhong, Yiqi and Wang, Dequan and Wang, Yanfeng and Chen, Siheng},
  year = {2024},
  month = apr,
  number = {arXiv:2401.13964},
  eprint = {2401.13964},
  primaryclass = {cs},
  publisher = {arXiv},
  doi = {10.48550/arXiv.2401.13964},
  urldate = {2024-12-05},
  archiveprefix = {arXiv},
  langid = {english}
}

@inproceedings{wangCoreCooperativeReconstruction2023,
  title = {Core: {{Cooperative Reconstruction}} for {{Multi-Agent Perception}}},
  shorttitle = {Core},
  booktitle = {2023 {{IEEE}}/{{CVF International Conference}} on {{Computer Vision}} ({{ICCV}})},
  author = {Wang, Binglu and Zhang, Lei and Wang, Zhaozhong and Zhao, Yongqiang and Zhou, Tianfei},
  year = {2023},
  month = oct,
  pages = {8676--8686},
  publisher = {IEEE},
  address = {Paris, France},
  doi = {10.1109/ICCV51070.2023.00800},
  urldate = {2024-12-05},
  copyright = {https://doi.org/10.15223/policy-029},
  isbn = {979-8-3503-0718-4},
  langid = {english}
}

@misc{wangV2VNetVehicletoVehicleCommunication2020,
  title = {{{V2VNet}}: {{Vehicle-to-Vehicle Communication}} for {{Joint Perception}} and {{Prediction}}},
  shorttitle = {{{V2VNet}}},
  author = {Wang, Tsun-Hsuan and Manivasagam, Sivabalan and Liang, Ming and Yang, Bin and Zeng, Wenyuan and Tu, James and Urtasun, Raquel},
  year = {2020},
  month = aug,
  number = {arXiv:2008.07519},
  eprint = {2008.07519},
  primaryclass = {cs},
  publisher = {arXiv},
  doi = {10.48550/arXiv.2008.07519},
  urldate = {2024-12-12},
  archiveprefix = {arXiv},
  langid = {english}
}

@misc{xuV2XViTVehicletoEverythingCooperative2022,
  title = {{{V2X-ViT}}: {{Vehicle-to-Everything Cooperative Perception}} with {{Vision Transformer}}},
  shorttitle = {{{V2X-ViT}}},
  author = {Xu, Runsheng and Xiang, Hao and Tu, Zhengzhong and Xia, Xin and Yang, Ming-Hsuan and Ma, Jiaqi},
  year = {2022},
  month = aug,
  number = {arXiv:2203.10638},
  eprint = {2203.10638},
  primaryclass = {cs},
  publisher = {arXiv},
  doi = {10.48550/arXiv.2203.10638},
  urldate = {2025-01-06},
  archiveprefix = {arXiv},
  langid = {english}
}

@article{yangHow2commCommunicationEfficientCollaborationPragmatic2023,
  title = {How2comm: {{Communication-Efficient}} and {{Collaboration-Pragmatic Multi-Agent Perception}}},
  author = {Yang, Dingkang and Yang, Kun and Wang, Yuzheng and Liu, Jing and Xu, Zhi and Yin, Rongbin and Zhai, Peng and Zhang, Lihua},
  year = {2023},
  langid = {english}
}

@article{arnold2020cooperative,
  title={Cooperative perception for 3D object detection in driving scenarios using infrastructure sensors},
  author={Arnold, Eduardo and Dianati, Mehrdad and De Temple, Robert and Fallah, Saber},
  journal={IEEE Transactions on Intelligent Transportation Systems},
  volume={23},
  number={3},
  pages={1852--1864},
  year={2020},
  publisher={IEEE}
}

@article{xu2025cosdh,
  title={CoSDH: Communication-Efficient Collaborative Perception via Supply-Demand Awareness and Intermediate-Late Hybridization},
  author={Xu, Junhao and Zhang, Yanan and Cai, Zhi and Huang, Di},
  journal={arXiv preprint arXiv:2503.03430},
  year={2025}
}

@inproceedings{codefilling,
  title={Communication-efficient collaborative perception via information filling with codebook},
  author={Hu, Yue and Peng, Juntong and Liu, Sifei and Ge, Junhao and Liu, Si and Chen, Siheng},
  booktitle={Proceedings of the IEEE/CVF Conference on Computer Vision and Pattern Recognition},
  pages={15481--15490},
  year={2024}
}

@inproceedings{opv2v,
author = {Xu, Runsheng and Xiang, Hao and Xia, Xin and Han, Xu and Li, Jinlong and Ma, Jiaqi},
title = {OPV2V: An Open Benchmark Dataset and Fusion Pipeline for Perception with Vehicle-to-Vehicle Communication},
year = {2022},
doi = {10.1109/ICRA46639.2022.9812038},
booktitle = {2022 International Conference on Robotics and Automation (ICRA)},
pages = {2583–2589},
numpages = {7},
location = {Philadelphia, PA, USA}
}

@inproceedings{dair-v2x,
  title={Dair-v2x: A large-scale dataset for vehicle-infrastructure cooperative 3d object detection},
  author={Yu, Haibao and Luo, Yizhen and Shu, Mao and Huo, Yiyi and Yang, Zebang and Shi, Yifeng and Guo, Zhenglong and Li, Hanyu and Hu, Xing and Yuan, Jirui and Nie, Zaiqing},
  booktitle={Proceedings of the IEEE/CVF Conference on Computer Vision and Pattern Recognition},
  pages={21361--21370},
  year={2022}
}

@article{V2X-R,
  title={V2X-R: Cooperative LiDAR-4D Radar Fusion for 3D Object Detection with Denoising Diffusion},
  author={Huang, Xun and Wang, Jinlong and Xia, Qiming and Chen, Siheng and Yang, Bisheng and Wang, Cheng and Wen, Chenglu},
  journal={arXiv preprint arXiv:2411.08402},
  year={2024}
}

@inproceedings{lang2019pointpillars,
  title={Pointpillars: Fast encoders for object detection from point clouds},
  author={Lang, Alex H and Vora, Sourabh and Caesar, Holger and Zhou, Lubing and Yang, Jiong and Beijbom, Oscar},
  booktitle={Proceedings of the IEEE/CVF conference on computer vision and pattern recognition},
  pages={12697--12705},
  year={2019}
}

@article{zhu2020deformable,
  title={Deformable detr: Deformable transformers for end-to-end object detection},
  author={Zhu, Xizhou and Su, Weijie and Lu, Lewei and Li, Bin and Wang, Xiaogang and Dai, Jifeng},
  journal={arXiv preprint arXiv:2010.04159},
  year={2020}
}

@inproceedings{yang2024align,
  title={Align before collaborate: Mitigating feature misalignment for robust multi-agent perception},
  author={Yang, Kun and Yang, Dingkang and Li, Ke and Xiao, Dongling and Shao, Zedian and Sun, Peng and Song, Liang},
  booktitle={European Conference on Computer Vision},
  pages={282--299},
  year={2024},
  organization={Springer}
}

@article{tao2024directed,
  title={Directed-CP: Directed Collaborative Perception for Connected and Autonomous Vehicles via Proactive Attention},
  author={Tao, Yihang and Hu, Senkang and Fang, Zhengru and Fang, Yuguang},
  journal={arXiv preprint arXiv:2409.08840},
  year={2024}
}

@article{wang2025coopdetr,
  title={CoopDETR: A unified cooperative perception framework for 3D detection via object query},
  author={Wang, Zhe and Xu, Shaocong and Zhuang, Xucai and Xu, Tongda and Wang, Yan and Liu, Jingjing and Chen, Yilun and Zhang, Ya-Qin},
  journal={arXiv preprint arXiv:2502.19313},
  year={2025}
}

@article{zhang2025co,
  title={Co-mtp: A cooperative trajectory prediction framework with multi-temporal fusion for autonomous driving},
  author={Zhang, Xinyu and Zhou, Zewei and Wang, Zhaoyi and Ji, Yangjie and Huang, Yanjun and Chen, Hong},
  journal={arXiv preprint arXiv:2502.16589},
  year={2025}
}

@article{lu2022robust,
  title={Robust collaborative 3d object detection in presence of pose errors},
  author={Lu, Yifan and Li, Quanhao and Liu, Baoan and Dianati, Mehrdad and Feng, Chen and Chen, Siheng and Wang, Yanfeng},
  journal={arXiv preprint arXiv:2211.07214},
  year={2022}
}

@inproceedings{shao2024hetecooper,
  title={Hetecooper: Feature collaboration graph for heterogeneous collaborative perception},
  author={Shao, Congzhang and Luo, Guiyang and Yuan, Quan and Chen, Yifu and Liu, Yilin and Gong, Kexin and Li, Jinglin},
  booktitle={European Conference on Computer Vision},
  pages={162--178},
  year={2024},
  organization={Springer}
}

@INPROCEEDINGS{dsrc,
  author={Bera, R. and Bera, J. and Sil, S. and Dogra, S. and Sinha, N.B. and Mondal, D.},
  booktitle={2006 IFIP International Conference on Wireless and Optical Communications Networks}, 
  title={Dedicated short range communications (DSRC) for intelligent transport system}, 
  year={2006},
  volume={},
  number={},
  pages={5 pp.-5},
  doi={10.1109/WOCN.2006.1666607}}

@article{xu2025codyntrust,
  title={CoDynTrust: Robust Asynchronous Collaborative Perception via Dynamic Feature Trust Modulus},
  author={Xu, Yunjiang and Li, Lingzhi and Wang, Jin and Yang, Benyuan and Wu, Zhiwen and Chen, Xinhong and Wang, Jianping},
  journal={arXiv preprint arXiv:2502.08169},
  year={2025}
}

\clearpage
\appendix

\subsection{Datasets}

\noindent\textbf{OPV2V}~\cite{opv2v}. OPV2V is one of the earliest simulation datasets designed for Vehicle-to-Vehicle (V2V) communication. In our experiments, we set the maximum number of collaborating vehicles to 5.

\noindent\textbf{V2X-R}~\cite{V2X-R}. V2X-R is the first simulation dataset for collaborative perception that incorporates the 4D radar modality. Similar to OPV2V, we limit the maximum number of collaborating vehicles to 5.

\noindent\textbf{DAIR-V2X}~\cite{dair-v2x}. DAIR-V2X is the first real-world Vehicle-to-Everything (V2X) dataset. It features one ego vehicle collaborating with infrastructure-mounted sensors located at road intersections. In our setup, the ego vehicle communicates with a single infrastructure unit, as the elevated position of infrastructure sensors provides a broader field of view compared to vehicle-mounted sensors.

\subsection{Training Configurations}

\noindent\textbf{Bandwidth Settings.} To ensure a fair comparison under an equal number of shared features (\ie sampling in the spatial plane of the BEV feature), we standardize the resolution of intermediate features within the same spatial coverage, as summarized in Table~\ref{tab:feature-resolution}. Nevertheless, the actual communication bandwidth still varies across methods due to additional design differences.  
\begin{table}[thb]\centering
    \caption{Perception range, downsampling rate, and BEV resolution for each dataset.}
    \label{tab:feature-resolution}
    \resizebox{0.48\textwidth}{!}{
    \large
    \begin{tabular}{c c c c}
        \toprule
        Dataset & \makecell{Perception \\ Range (Meters)} & \makecell{Downsampling \\ Rate} & \makecell{BEV \\ Resolution} \\
        \midrule
        OPV2V    & $[-140.8, -38.4, -3,\; 140.8, 38.4, 1]$   & 4 & $176 \times 48$ \\
        V2X-R    & $[-140.8, -38.4, -3,\; 140.8, 38.4, 1]$   & 4 & $176 \times 48$ \\
        DAIR-V2X & $[-100.8, -40, -3.5,\; 100.8, 40, 1.5]$   & 2 & $126 \times 50$ \\
        \bottomrule
    \end{tabular}
    }
\end{table}

For \textit{dense sharing methods} (\ie V2VNet, When2Comm, V2XViT, and AttFuse), we disable channel compression and quantization to ensure consistency. Notably, When2Comm applies agent pruning, which reduces its bandwidth usage compared to the other three dense sharing baselines:
\[
\text{V2VNet} = \text{V2XViT} =  \text{AttFuse} >  \text{When2Comm}.
\]

For \textit{sparse sharing methods} (\ie Where2Comm and CORE), only a subset of BEV feature grids is sampled in the spatial plane according to a predefined ratio. For CoSDH and our proposed \framework, we adopt an additional \emph{channel downsampling} strategy, reducing the channel dimension of each BEV grid from $C=256$ to $C'=16$, combined with FP16 quantization where each value is represented with 16-bit floating point precision. As a result, although CoSDH and \framework use the same selection ratio in the BEV plane as Where2Comm and CORE, their overall bandwidth consumption is significantly lower, following the relation:  
\[
\text{Where2Comm} = \text{CORE} > \text{CoSDH} = \text{\framework}.
\]

\noindent\textbf{Loss Function.} For OPV2V and DAIR datasets, following prior work, we adopt a multi-task detection loss that combines classification, regression, and direction supervision. Specifically, the classification branch employs a Sigmoid Focal Loss with $\alpha = 0.25$ and $\gamma = 2.0$, and we set the positive classification weight to $2.0$ to address class imbalance. The regression branch is optimized with a Weighted Smooth L1 Loss ($\sigma = 3.0$) applied in a codewise manner, with weight $2.0$. To further improve orientation estimation, we incorporate a direction classification branch optimized by a Weighted Softmax Loss with weight $0.2$. The overall loss is  
\[
\mathcal{L}_{\text{OPV2V/DAIR}} 
= \mathcal{L}_{\text{cls}} + \mathcal{L}_{\text{reg}} + \mathcal{L}_{\text{dir}} .
\]

For V2X-R, we adopt a simpler detection loss consisting only of classification and regression terms as in its original codebase:  
\[
\mathcal{L}_{\text{V2X-R}} 
= \mathcal{L}_{\text{cls}} + 2.0 \cdot \mathcal{L}_{\text{reg}} ,
\]
with classification weight $1.0$ and regression weight $2.0$. 

For CORE, in addition to classification and regression, it explicitly supervises the reconstruction of representation with an MSE loss,  
\[
\mathcal{L}_{\text{CORE}} 
= \mathcal{L}_{\text{cls}} + \mathcal{L}_{\text{reg}} + \mathcal{L}_{\text{rec}}, 
\quad 
\mathcal{L}_{\text{rec}} = \|\mathbf{x}_{\text{rec}} - \mathbf{x}_{\text{ideal}}\|_2^2 .
\]

Finally, for our proposed framework, beyond the OPV2V/DAIR/V2X-R detection loss, we explicitly supervise the foreground prediction head with a binary cross-entropy loss on the center mask, resulting in  
\[
\mathcal{L}_{\text{Ours}} 
= \mathcal{L}_{\text{cls}} + \mathcal{L}_{\text{reg}} + (\mathcal{L}_{\text{dir}}) + \mathcal{L}_{\text{ctr}} ,
\]
where $\mathcal{L}_{\text{ctr}}$ encourages accurate foreground localization.

\noindent\textbf{Learning Rate and Optimizer.}  
For OPV2V, we employ the Adam optimizer with a learning rate of $2\times 10^{-3}$, $\epsilon = 1 \times 10^{-10}$, and weight decay $1\times 10^{-4}$. The learning rate is scheduled using a MultiStep policy with decay factor $\gamma = 0.1$ and milestones at epochs 10 and 20. For DAIR, we adopt the same optimizer settings but with a lower initial learning rate of $1\times 10^{-3}$ while keeping the same scheduler configuration. For V2X-R, we again use Adam with learning rate $2\times 10^{-3}$, $\epsilon = 1 \times 10^{-10}$, and weight decay $1\times 10^{-4}$, while the MultiStep scheduler decays the learning rate by $\gamma = 0.1$ at epochs 10 and 15. In some cases (\eg AttFuse, etc), we alternatively employ a cosine annealing scheduler with warmup, where the learning rate is linearly increased from $2\times 10^{-5}$ to the base learning rate during the first 10 epochs and then decayed to a minimum of $5\times 10^{-6}$ over the course of training. These configurations are our main setup across datasets. However, we adapt hyperparameters when necessary to reproduce the best performance reported in prior works to ensure a fair comparison. For full setup details, please refer to our released code and configurations.

\begin{figure*}[t!] \centering
    \includegraphics[width=\textwidth]{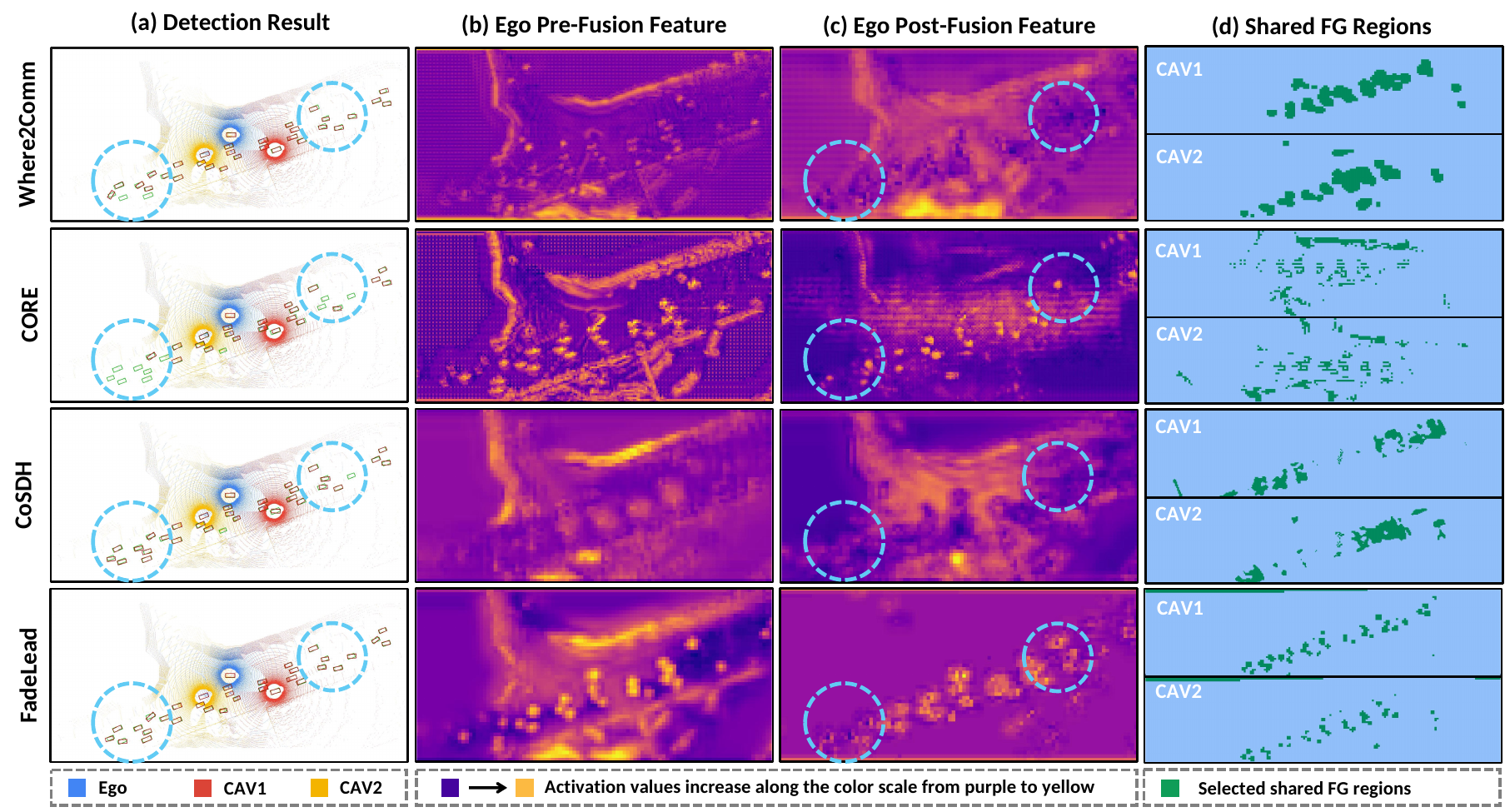}
    \caption{Visualization on \textbf{OPV2V}~\cite{opv2v} with \textbf{top-5\%} confident foreground selection. We show detection results, the ego BEV feature before fusion, the ego BEV feature after fusion, and the shared regions from collaborating CAVs.} \label{fig:opv2v_tk005_vis}
    \vspace{-1em}
\end{figure*}

\begin{figure*}[t!] \centering
    \includegraphics[width=\textwidth]{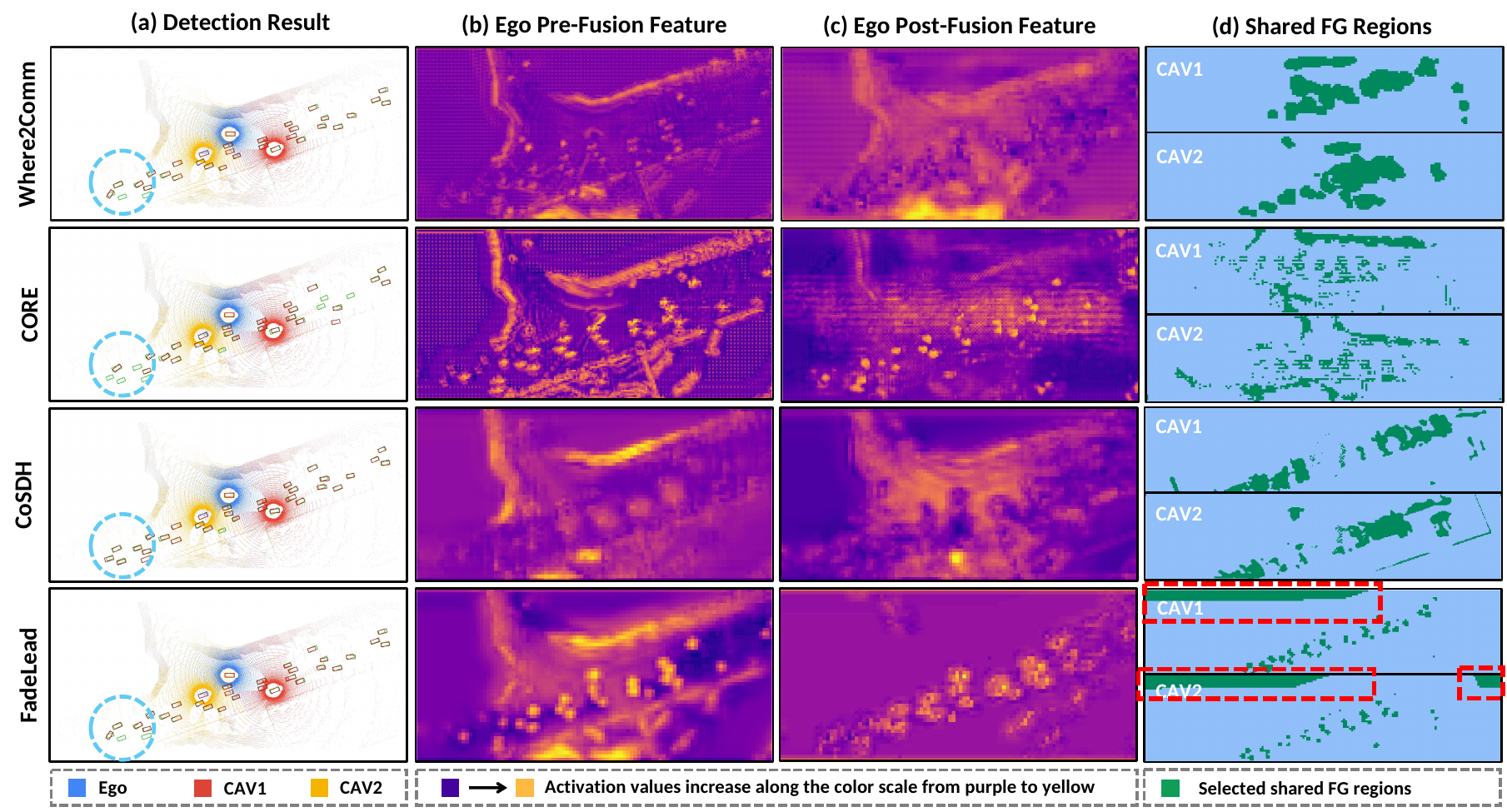}
    \caption{Visualization on \textbf{OPV2V}~\cite{opv2v} with \textbf{top-10\%} confident foreground selection. We show detection results, the ego BEV feature before fusion, the ego BEV feature after fusion, and the shared regions from collaborating CAVs.} \label{fig:opv2v_tk01_vis}
    \vspace{-1em}
\end{figure*}

\subsection{Further Results}

\noindent\textbf{Visualizations on OPV2V of top-$5\%$.} As illustrated in Fig.~\ref{fig:opv2v_tk005_vis}, \framework consistently highlights foreground regions that closely align with the ground-truth objects. After fusion, the post-fusion BEV feature maps remain cleanly separated between foreground and background, preserving object structures. Compared to the 1\% setting, increasing the sharing ratio to 5\% notably improves the reconstructed post-fusion features for CORE, reducing noise and sharpening object activations. By contrast, CoSDH and Where2Comm still fail to capture several objects, as indicated by missing high activations in the corresponding regions of their post-fusion maps. This demonstrates that \framework maintains robustness even under stricter selection budgets, while alternative methods suffer from incomplete context sharing.

\noindent\textbf{Visualization on OPV2V at Top-$10\%$.} As shown in Fig.~\ref{fig:opv2v_tk01_vis}, increasing the sharing ratio to $10\%$ improves recall for CORE and Where2Comm. In contrast, CoSDH, guided by point cloud density, tends to oversample drivable regions and introduces noise, while our \framework maintains a compact foreground mask but includes boundary regions. Such additional background can even degrade performance. These results suggest that \framework has already absorbed sufficient background cues into its enriched foreground, and that further gains arise from selectively transmitting high-confidence regions rather than simply enlarging the top-$k$ set. For deployment, we therefore recommend a \emph{threshold-based selection} strategy over fixed top-$k$, ensuring robust and bandwidth-efficient collaboration across diverse scenes.

\noindent\textbf{Visualizations on V2X-R/DAIR-V2X of top-$1\%/5\%/10\%$.} Please refer to our uploaded visualization videos.

\subsection{Details of Ablation Study}
We conduct ablation experiments on the three core modules of our framework—FCA, CBP, and FAF—to validate their contributions.  

\noindent\textbf{Without FCA.}  
In the full method, Foreground Context Attention refines the foreground representation by incorporating density priors and applying deformable attention:
\[
\tilde{\mathbf{F}}^{\text{FG}}_i = \mathrm{Attn}_{\text{deform}}\!\big(\mathbf{F}^{\text{BEV}}_i,\, \mathbf{C}'_i\big),
\]
where $\mathbf{C}'_i$ is the density-refined confidence map.  
In the ablated variant, we bypass FCA and directly use the raw foreground feature from $\mathbf{F}^{\text{BEV}}_i$:  
\[
\tilde{\mathbf{F}}^{\text{FG, w/o FCA}}_i = \mathbf{F}^{\text{FG}}_i .
\]
This comparison highlights the effect of contextual enrichment through deformable attention.

\noindent\textbf{Without CBP.}  
Our CBP module progressively selects informative background regions $\mathbf{F}^{\text{BG-Info}}_i$ during training, yielding the transmitted feature:
\[
\mathbf{F}^{\text{sh}}_i = \tilde{\mathbf{F}}^{\text{FG}}_i \cup \mathbf{F}^{\text{BG-Info}}_i .
\]
As the training curriculum advances, the ratio of $\mathbf{F}^{\text{BG-Info}}_i$ is annealed to zero, leaving only foreground transmission at inference.  
In the ablation, we remove CBP entirely and force background exclusion from the start, transmitting only the top-$k$ predicted foreground patches:
\[
\mathbf{F}^{\text{sh, w/o CBP}}_i = \text{TopK}_k(\mathbf{F}^{\text{FG}}_i).
\]
This stresses the importance of curricular background pruning in absorbing contextual cues into the foreground.

\noindent\textbf{Without FAF.}  
In our framework, features received from neighbors $\mathcal{N}(i)$ are integrated with ego features via context-aware fusion:
\[
\mathbf{F}^{\text{fused}}_i = \mathbf{F}^{\text{ego}}_i + \mathrm{FAF}\!\big(\{\mathbf{F}^{\text{sh}}_j, \mathbf{M}^{\text{sh}}_j\}_{j \in \mathcal{N}(i)}\big).
\]
FAF gates updates by the transmission mask $\mathbf{M}^{\text{sh}}_j$ and aligns distributions before fusion.  
In the ablation, we replace FAF with element-wise maximum fusion across collaborators:
\[
\mathbf{F}^{\text{fused, w/o FAF}}_i = \max_{j \in \mathcal{N}(i)} \mathbf{F}^{\text{sh}}_j,
\]
These ablations collectively demonstrate that FCA provides refined and context-aware foregrounds, CBP enables gradual absorption of background semantics into transmitted features, and FAF ensures robust and distributionally aligned multi-agent fusion. Their removal leads to performance degradation, verifying that each module contributes to the robustness and efficiency of the full framework.

\balance

\end{document}